\documentclass{article}

\usepackage[english]{babel}

\usepackage[letterpaper,top=2cm,bottom=2cm,left=3cm,right=3cm,marginparwidth=1.75cm]{geometry}

\usepackage{amsmath}
\usepackage{graphicx}
\usepackage[colorlinks=true, allcolors=blue]{hyperref}
\usepackage{cite}
\usepackage{amsmath, amssymb}
\usepackage{geometry}
\usepackage{xcolor}
\usepackage{enumitem}
\usepackage{listings}
\usepackage{xcolor}
\usepackage{listings}
\usepackage{tcolorbox}
\tcbuselibrary{skins, breakable, listings}

\newtcolorbox{promptbox}[1]{
	colback=gray!5,
	colframe=blue!50!black,
	fonttitle=\bfseries,
	title=#1,
	enhanced,
	attach boxed title to top left={xshift=5mm, yshift=-2mm},
	boxed title style={size=small, colback=blue!50!black, colframe=blue!50!black},
	breakable,
	listing engine=listings,
	listing options={
		basicstyle=\ttfamily\small,
		breaklines=true,
		breakatwhitespace=true,
		showstringspaces=false,
		tabsize=2,
		keywordstyle={},
		commentstyle={},
		stringstyle={}
	}
}
\usepackage{authblk}

\title{Can LLM generate interesting mathematical research problems?}
\author{Xiaoyang Chen}
\author{Xiang Jiang}
\affil{School of Mathematical Science , Tongji University , China}

\begin{document}
\maketitle

\begin{abstract}
This paper is the second one in a series of work on the mathematical creativity of LLM. In the first paper, the authors proposed three criteria for evaluating the mathematical creativity of LLM and constructed a benchmark dataset to measure it. This paper further explores the mathematical creativity of LLM, with a focus on investigating whether LLM can generate valuable and cutting-edge mathematical research problems. We develop an agent to generate unknown problems and produced 665 research problems in differential geometry. Through human verification, we find that many of these mathematical problems are unknown to experts and possess unique research value.

\end{abstract}

\section{Introduction}

The current research on the application of large language models in mathematics is experiencing explosive growth. Mainstream evaluation benchmarks and model optimization directions are almost exclusively focused on reasoning ability—that is, whether a model can derive the correct answer step by step to solve mathematical problems. However, mathematics is more than just a mechanical combination of logical derivations. The soul of mathematics lies in creativity—proposing unprecedented concepts, inventing ingenious methods, and constructing counterexamples that subvert established understanding.

This paper is the second in a series of work on the mathematical creativity of LLM. In the first paper \cite{chen2025deepmath}, to scientifically evaluate creativity, the authors proposed an evaluation model based on three key criteria:
\begin{enumerate}
\item Generation of New Concepts  
This represents the highest level of creativity, reflected in the introduction of unprecedented mathematical concepts or ideas that open up entirely new fields of research.  For example, the emergence of the Riemannian metric laid the foundation for modern differential geometry and ultimately provided the mathematical language for Einstein's theory of general relativity.  

\item Invention of New Methods 
Another form of creativity involves inventing entirely new techniques or tools to solve previously intractable complex problems.  
For instance, a milestone method in geometric analysis is the Bochner technique, which ingeniously connects geometry with analysis, using curvature conditions to constrain the topological properties of manifolds such as Betti numbers, offering an elegant pathway for proving many geometric theorems.

\item Creation of New Mathematical Objects
In the process of mathematical research, it is often necessary to construct specific mathematical objects. For example, to obtain a priori estimation of partial differential equations, one typically needs to construct special auxiliary functions. On the other hand, to prove that certain mathematical propositions are false, mathematicians need to construct specific counterexamples.  In contrast to abstract concepts and theories, these concrete, constructible mathematical objects are more amenable to verification and thus offer a valuable entry point for evaluating the creative capabilities of large language models. 
\end{enumerate}

In the subsequent research work, based on the above three evaluative dimensions of creativity, we will study the mathematical creativity of LLM from various perspectives. In this paper, we focus on exploring whether LLM can generate valuable, cutting-edge mathematical research questions. We build an agent to generate unknown problems and produced 665 problems in differential geometry. Through human verification, we find that many of these mathematical problems are unknown to experts and possess unique research value. In future work, we plan to build on this dataset by incorporating reinforcement learning to enhance the mathematical capabilities of LLM for creative thinking.

\section{DeepMath-generate: an agent to generate research problems}

Proposing questions and solving problems are two distinct aspects of mathematical research. While solving problems is certainly important, asking a good question is equally crucial, as a good question can lead new research directions. Current mathematical research on LLMs focuses on whether LLM can accurately solve certain challenging mathematical problems.

For example, mathematicians from top research institutions recently released a dataset named "first proof" to test the ability of LLM to solve research-level mathematical problems \cite{abouzaid2026first}. Gemini Deep Think managed to solve 6 out of the 10 problems in "first proof"\cite{feng2026aletheia}. Although these problems are extremely difficult, they all have standard answers.

This paper focuses on investigating whether LLM can generate valuable mathematical research questions. First, we need to address a fundamental question: What makes a good mathematical problem?

In mathematics, judging whether a problem is a "good problem" depends not merely on its difficulty, but on whether it can advance the development of the field, reveal deep structures, or connect seemingly unrelated domains. Based on the characteristics of cutting-edge mathematical research, a good mathematical problem typically possesses the following core attributes:
\begin{enumerate}
\item Profound Insight and Foresight
A good problem often directly targets the core essence of a discipline or anticipates a potential new field.
For example, the Riemann Hypothesis, proposed by Bernhard Riemann in 1859, was not an arbitrary computational exercise but a profound insight into the distribution of prime numbers. Although it remains unproven to this day, it has spurred the development of multiple branches, such as analytic number theory and algebraic geometry, over the past century and beyond. A good problem is, in itself, a beacon.

\item Serving as a Bridge Between Different Fields
A key characteristic of cutting-edge mathematics is cross-disciplinary integration. Problems that can connect two seemingly unrelated fields are often significant.
For instance, the Atiyah–Singer Index Theorem profoundly reveals the intrinsic link between analysis (the number of solutions to differential equations) and topology (the geometric shape of manifolds). If a mathematical problem compels one to invoke tools from another field to solve it, it often leads to revolutionary progress.

\item Extreme Simplicity
A beautiful problem is often expressed in extremely simple terms, sometimes understandable even to non-experts.
Consider the Poincaré Conjecture. Its statement is remarkably straightforward: "Every simply connected, closed 3-manifold is diffeomorphic to a 3-sphere?" Anyone with a basic understanding of topological concepts can grasp this statement. Yet it baffled topologists for an entire century until Grigori Perelman solved it using Ricci flow. This kind of problem—"narrow entrance, profound depth"—represents the pinnacle of mathematical aesthetics.
\end{enumerate}

To enable LLM to generate good mathematical problems, we construct a specialized agent: DeepMath-generate. This agent consists of two components: a generator, which creates mathematical problems, and an evaluator, which assesses whether a generated problem is "valuable."

The generator creates mathematical problems based on the instructions in the system prompt, and then the evaluator assesses the generated problems according to the prompt’s requirements. Subsequently, the generator produces improved mathematical problems based on the evaluator's feedback, and this iterative process continues. The generator and the evaluator utilize the GPT 5 / GPT 5.3 model via API calls respectively, and the detailed code is available on the GitHub homepage:\url{https://github.com/DeepMathLLM/DeepMath/tree/main/DeepMath-generator}..  

To achieve good results, appropriate prompts are crucial. The core of the entire process lies in how to make the LLM understand the essence of a good mathematical problem. On the other hand, since a LLM has learned a vast amount of mathematical knowledge, it is necessary to design a basic filtering mechanism to prevent the generated mathematical problems from overlapping with existing mathematical research. Based on these considerations, the complete system prompt we design is as follows:

\begin{promptbox}{Generator Prompt}
	You are a mathematical problem generator specializing in "good" research-level mathematics problems.
	
	Follow the instructions strictly.
	
	**Task**
	
	Generate one mathematics problem based on the knowledge points provided by the user.
	
	**Requirements**
	
	1. Do not generate problems that are simply the statement or conclusion of an existing theorem.
	
	In particular, do not restate well-known theorems or their standard conclusions as problem content (for example, do not generate problems that essentially ask to prove a classical theorem directly).
	
	2. Each problem must be at a research level and the answers are still unknown.
	Each problem must be at a level suitable for senior mathematicians conducting original research and not a known theorem or standard exercise.
	
	3. Each problem should be a "good" problem. Here a good mathematical problem typically possesses the following core attributes:
	
	- **Profound Insight and Foresight**
	
	A good problem often directly targets the core essence of a discipline or anticipates a potential new field.
	
	For example, the Riemann Hypothesis, proposed by Bernhard Riemann in 1859, was not an arbitrary computational exercise but a profound insight into the distribution of prime numbers. Although it remains unproven to this day, it has spurred the development of multiple branches, such as analytic number theory and algebraic geometry, over the past century and beyond. A good problem is, in itself, a beacon.
	
	- **Serving as a Bridge Between Different Fields**
	
	A key characteristic of cutting-edge mathematics is cross-disciplinary integration. Problems that can connect two seemingly unrelated fields are often good and significant. For instance, the Atiyah–Singer Index Theorem profoundly reveals the intrinsic link between analysis (the number of solutions to differential equations) and topology (the geometric shape of manifolds). If a mathematical problem compels one to invoke tools from another field to solve it, it often leads to revolutionary progress.
	
	- **Extreme Simplicity**
	
	A good and beautiful problem is often expressed in extremely simple terms, sometimes understandable even to non-experts.
	Consider the Poincaré Conjecture. Its statement is remarkably straightforward: "Every simply connected, closed 3-manifold is diffeomorphic to a 3-sphere?" Anyone with a basic understanding of topological concepts can grasp this statement. Yet it baffled topologists for an entire century until Grigori Perelman solved it using Ricci flow. This kind of problem—"narrow entrance, profound depth"—represents the pinnacle of mathematical aesthetics.
	
	4. Problems must be complex and non-trivial.
	
	Avoid straightforward or routine exercises.
	
	5. Ensure that every problem is logically valid.
	
	Do not generate problems that contain contradictions, impossible conditions, undefined objects, or statements that are clearly false or ill-posed. The problem itself must be mathematically sound.
	
	6. After the problem, you must include the following explanation:
	
	Why it is a "good" problem:
	
	Explain the deep reason why it serves as a "good" problem.
	
	**Output Format (must be followed exactly)**
	
	problem:
	
	(Problem content)
	
	Why is it a "good" problem:
	
	**Revision Rules**
	
	- If the user provides revision feedback, you must modify the problem accordingly.
	After each revision, you must output the complete problem again, not only the modified ones.
	
	- Do not omit any problem or explanation section.
	
	- Always maintain exactly the same output format.
	
\end{promptbox}

\begin{promptbox}{Evaluator Prompt}
	You are a mathematical problem evaluator whose role is to critically assess whether the given problems genuinely qualifies as "good" research-level mathematics problems and whether their explanations are logically sound and sufficiently justified.
	
	You must not generate new problems. Your task is only to evaluate the given problems and their explanations.
	
	**Evaluation Task**
	
	For the provided problem, carefully evaluate the following aspects:
	
	1. Check for Existing Theorem Statements
	
	Determine whether the problem is simply the statement or conclusion of a known theorem.
	If the problem essentially restates a classical theorem or directly asks for its proof, it does not meet the requirement and must be flagged.
	
	2. Research level:
	
	The problem must be at a level suitable for senior mathematicians conducting original research. It requires deep insight, advanced techniques, or novel connections beyond what is typically encountered in graduate coursework or standard literature.
	
	3. Unknown answer
	
	The problem must be not a known theorem or standard exercise; its answer is not established in existing literature.
	
	4. Quality of the explanation
	
	Assess whether the explanation clearly and convincingly justifies why the problem qualifies as a "good" problem.
	
	The reasoning should explicitly connect the structure of the problem to the standards of a strong mathematical research problem.
	
	5. Check whether the problem is a "good" problem.
	
	Here a good mathematical problem typically possesses the following core attributes:
	
	- **Profound Insight and Foresight**
	
	A good problem often directly targets the core essence of a discipline or anticipates a potential new field.
	
	For example, the Riemann Hypothesis, proposed by Bernhard Riemann in 1859, was not an arbitrary computational exercise but a profound insight into the distribution of prime numbers. Although it remains unproven to this day, it has spurred the development of multiple branches, such as analytic number theory and algebraic geometry, over the past century and beyond. A good problem is, in itself, a beacon.
	
	- **Serving as a Bridge Between Different Fields**
	
	A key characteristic of cutting-edge mathematics is cross-disciplinary integration. Problems that can connect two seemingly unrelated fields are often good and significant. For instance, the Atiyah–Singer Index Theorem profoundly reveals the intrinsic link between analysis (the number of solutions to differential equations) and topology (the geometric shape of manifolds). If a mathematical problem compels one to invoke tools from another field to solve it, it often leads to revolutionary progress.
	
	- **Extreme Simplicity**
	
	A good and beautiful problem is often expressed in extremely simple terms, sometimes understandable even to non-experts.
	
	Consider the Poincaré Conjecture. Its statement is remarkably straightforward: "Every simply connected, closed 3-manifold is diffeomorphic to a 3-sphere?" Anyone with a basic understanding of topological concepts can grasp this statement. Yet it baffled topologists for an entire century until Grigori Perelman solved it using Ricci flow. This kind of problem—"narrow entrance, profound depth"—represents the pinnacle of mathematical aesthetics.
	
	6. Logical Validity of the Problem
	
	Verify that the problem itself is mathematically sound.
	Identify whether there are:
	
	- contradictions in the conditions
	
	- impossible requirements
	
	- undefined objects or ambiguous statements
	
	- claims that are clearly false
	
	If such issues exist, clearly explain the logical flaw.
	
	**Evaluation Output Rules**
	
	For the problem:
	
	-If the explanations are vague, superficial, incorrect, or logically insufficient, identify the specific issue.
	
	-If the problem is simply a restatement of an existing theorem, explicitly point this out.
	
	-If the problem contains logical flaws, clearly explain the issue.
	
	When necessary, provide specific revision suggestions, including:
	
	- how to improve the research depth of the problem
	
	- how to improve the clarity and rigor of the explanation
	
	- how to better align the problem with the standards of a "good" mathematical problem
	
	Your feedback should be precise, critical, and mathematically rigorous.
	
	If the problem and its explanation are fully valid, logically rigorous, truly research-level in nature, and are not merely statements or conclusions of existing theorems, and contain no obvious logical problems, then you should re-output the SAME problem statement and its relative explanation wrapped in <problem></problem> tags. And output the single word “Terminate” on a separate final line.
	
\end{promptbox}

\section{Results}

After designing the basic architecture of the agent, we applied it to specific research directions in differential geometry. We select 200 different research directions including  harmonic maps, curvature and topology, etc (see the appendix for a complete description), and use the agent to generate five distinct research problems for each specific direction. Through human verification, we find that many of these mathematical problems are unknown to experts and possess unique research value. Here are some concrete examples. 

problem 1:  

Let $\Sigma^k$ be an exotic $k$‑sphere ($k \ge 7$). For a fixed integer $r > 0$, consider a smooth rank‑$r$ vector bundle $E \to \Sigma^k$ and suppose its total space admits a complete Riemannian metric $g$ with nonnegative sectional curvature. The Cheeger–Gromoll Soul Theorem ensures that the soul $S$ of $(E,g)$ is diffeomorphic to $\Sigma^k$. 

For which pairs $(\Sigma^k, r)$ can such a metric exist? In particular, does there exist a positive integer $r$ such that some nontrivial rank‑$r$ vector bundle over an exotic sphere $\Sigma^k$ admits a complete nonnegatively curved metric whose soul $S$ is diffeomorphic to $\Sigma^k$?  

Equivalently, can exotic smooth structures on $S^k$ occur as the soul of a complete, nonnegatively curved manifold of codimension $r$, and if so, does there exist a minimal such $r$?  

Why is it a "good" problem:  

This question is mathematically consistent and precisely targets the interplay between nonnegative curvature and exotic smooth structures. It formalizes, in a sharply posed way, a geometric obstruction problem: can an exotic structure on a sphere survive as a soul under nonnegative curvature? By fixing the codimension $r$, the problem captures a progression that bridges global Riemannian geometry(through curvature and the soul theorem) and differential topology (through exotic spheres and vector bundle classifications).  

Its difficulty lies in the subtle curvature constructions on vector bundles (studied by Grove–Ziller, Belegradek–Kapovitch, and others) and the sensitivity of exotic smooth structures to bundle dimensions. Resolving even a single instance—positive or negative—would shed profound light on how curvature constrains smooth structures and could redefine our understanding of the geometric detectability of exotic differentiable types. The problem is simple to state, deeply open, and sits precisely at the confluence of curvature, topology, and smooth classification theory.

Problem 2: 

Let $\Sigma^k$ be an exotic smooth sphere of dimension $k \ge 7$.  

Denote by $\mathcal{R}_{\ge0}(\Sigma^k)$ the space of all smooth Riemannian metrics on $\Sigma^k$ with nonnegative sectional curvature, equipped with the $C^\infty$ topology, and let  
\[
\mathcal{M}_{\ge0}(\Sigma^k) = \mathcal{R}_{\ge0}(\Sigma^k) / \mathrm{Diff}_0(\Sigma^k)
\]
be the corresponding moduli space of nonnegatively curved metrics modulo diffeomorphisms isotopic to the identity.  

Similarly define $\mathcal{M}_{\ge0}(S^k)$ for the standard sphere.

Does there exist a dimension $k \ge 7$ and an exotic sphere $\Sigma^k$ such that $\mathcal{M}_{\ge0}(\Sigma^k)$ and $\mathcal{M}_{\ge0}(S^k)$ are *not homotopy equivalent?  Equivalently, can the exotic smooth structure on $\Sigma^k$ be detected by any topological invariant of the moduli space of nonnegatively curved metrics (for example, the number of connected components, $\pi_1$, or higher homotopy groups)?  

In particular, is it possible that $\mathcal{M}_{\ge0}(\Sigma^k)$ has more connected components than $\mathcal{M}_{\ge0}(S^k)$, reflecting geometrically distinct classes of nonnegatively curved metrics arising from the exotic differential structure?  

Or, conversely, is the smooth exoticity of $\Sigma^k$ completely invisible within the topology of its curvature moduli space?  

Why is it a "good" problem:  

This version is sharply formulated and mathematically sound: it fixes a precise moduli space $\mathcal{M}_{\ge0}$ (quotienting by $\mathrm{Diff}_0$) and asks whether exotic smooth structures can alter its fundamental topological features. It directly probes the detectability of smooth structure through curvature‑restricted moduli spaces, a question that lies squarely at the frontier of current Riemannian and differential topology.  

Resolving it would require deep insight into how nonnegative curvature interacts with the topology of diffeomorphism groups and smoothing theory. A positive answer would reveal that exotic smoothness leaves measurable traces in geometric moduli, while a negative answer would imply a surprising rigidity of nonnegative curvature under smooth variation. Either outcome would open new territory between infinite‑dimensional moduli theory, curvature geometry, and surgery‑theoretic classification of manifolds, unifying ideas from these distinct domains.  

The problem is also commendably simple to state—comparing two moduli spaces associated with homeomorphic manifolds—but is profound in depth: any progress would require breakthroughs in our understanding of the topology of spaces of nonnegatively curved metrics, making it a genuinely “good” and forward‑looking research problem.

The above two problems deal with the Riemannian geometry of exotic spheres, which is a widely studied topic. However, as far as the authors know, no one has studied Problem 1 and 2. The conclusions to be proved in these problems are deep and highly nontrivial, while solving these problems requires an extremely profound understanding.

\section{Discussion}
In this paper we develop an agent to generate unknown problems and produced 665 research problems in differential geometry. Through human verification, we find that many of these mathematical problems are unknown to experts and possess unique research value. However, these mathematical problems are not so exciting. For instance, the agent does not generate mathematical questions akin to the Poincaré conjecture. This requires further in-depth research, such as improving the prompts, focusing on more specific knowledge points, and refining the design of the agent. For more detailed information on the agent's design, please refer to \cite{luo2025large}.

\bibliographystyle{alpha}

\begin{thebibliography}{99}
	
	\bibitem{abouzaid2026first}
	M. Abouzaid et al.
	\newblock First Proof.
	\newblock \emph{arXiv preprint} arXiv:2602.05192, 2026.
	
	\bibitem{chen2025deepmath}
	X. Chen et al.
	\newblock DeepMath-Creative: A Benchmark for Evaluating Mathematical Creativity of Large Language Models.
	\newblock \emph{arXiv preprint} arXiv:2505.08744, 2025.
	
	\bibitem{feng2026aletheia}
	T. Feng et al.
	\newblock Aletheia tackles FirstProof autonomously.
	\newblock \emph{arXiv preprint} arXiv:2602.21201v2, 2026.
	
	\bibitem{luo2025large}
	J. Luo et al.
	\newblock Large Language Model Agent: A Survey on Methodology, Applications and Challenges.
	\newblock \emph{arXiv preprint} arXiv:2503.21460, 2025.
	
\end{thebibliography}

\section{Appendix}

\subsection*{200 Research Directions in Differential Geometry}

\paragraph{1. Minimal Submanifolds in Spheres and Space Forms}

\begin{enumerate}
	\item Existence and multiplicity of embedded minimal hyperspheres in $S^{n+1}$ with prescribed symmetries
	\item Free boundary minimal submanifolds in geodesic balls of the sphere 
	\item Index estimates for minimal hypersurfaces in spheres with lower Ricci curvature bounds
	\item Generic regularity of area-minimizing hypersurfaces in round spheres
	\item Morse theory for minimal submanifolds in space forms and applications to broken symmetries 
	\item Compactness theorems for minimal submanifolds in spheres with bounded second fundamental form
	\item Construction of new minimal surfaces in $S^3$ via gluing methods
	\item Stability of minimal submanifolds in Euclidean spheres under conformal deformations
	\item Gap phenomena for minimal submanifolds in spheres with positive curvature
	\item Minimal submanifolds with parallel second fundamental form in space forms
	\item Biharmonic submanifolds in Euclidean spheres and gap results 
	\item V-minimal submanifolds and their classification in space forms 
	\item Reduction techniques for minimal submanifolds from harmonic morphisms 
	\item Minimal Lagrangian submanifolds in complex space forms
	\item Willmore-type problems for minimal surfaces in spheres
	\item Asymptotic behavior of minimal submanifolds in hyperbolic space
	\item Minimal surfaces with prescribed boundary on the sphere
	\item Equivariant minimal submanifolds in spheres under cohomogeneity one actions
	\item Singularities of minimal varifolds in round spheres
	\item Minimal submanifolds and eigenvalue problems on the sphere
\end{enumerate}

\paragraph{2. Harmonic Maps and Their Generalizations}

\begin{enumerate}
	\setcounter{enumi}{20}
	\item Eells-Sampson type rigidity theorems under positive sectional curvature upper bounds 
	\item Harmonic maps from nonpositive curvature manifolds to positive curvature targets
	\item Sharp vanishing theorems for harmonic maps between manifolds with curvature constraints 
	\item Harmonic-Einstein metrics and their classification on closed manifolds 
	\item Stability of harmonic maps from Kähler manifolds to negatively curved targets
	\item Rigidity of harmonic maps from manifolds with nonnegative Ricci curvature
	\item Existence of harmonic maps between singular spaces with curvature bounds
	\item Harmonic maps from compact Kähler manifolds with positive scalar curvature 
	\item Vanishing theorems for harmonic maps to Kähler manifolds of strongly seminegative curvature 
	\item Heat flow methods for harmonic maps with singular targets
	\item Biharmonic maps and submanifolds in Sasakian space forms 
	\item Generalized harmonic maps and their applications in discrete geometry 
	\item Harmonic morphisms and their relation to minimal submanifolds 
	\item Vanishing theorems for harmonic functions on complete manifolds 
	\item Wave maps from Lorentzian manifolds to Riemannian targets
	\item Teichmüller theory and harmonic maps between surfaces
	\item Harmonic maps into buildings and rigidity theorems
	\item Approximately harmonic maps and quantitative rigidity estimates
	\item Harmonic map flow with critical points and bubble analysis
	\item Polynomial growth harmonic functions on manifolds with nonnegative curvature
\end{enumerate}

\paragraph{3. Manifolds with Nonnegative Sectional Curvature}

\begin{enumerate}
	\setcounter{enumi}{40}
	\item Topological classification of manifolds with nonnegative sectional curvature and torus actions 
	\item Almost isotropy-maximal manifolds with nonnegative curvature 
	\item Structure of noncompact manifolds with nonnegative sectional curvature 
	\item Splitting theorems for manifolds with nonnegative curvature and lines
	\item Fundamental groups of manifolds with nonnegative sectional curvature
	\item Betti number estimates for nonnegatively curved manifolds
	\item Rigidity of manifolds with nonnegative curvature and maximal symmetry rank
	\item Nonnegatively curved 4-manifolds with circle symmetry 
	\item Fixed point homogeneous manifolds with nonnegative curvature 
	\item Nonnegatively curved 5-manifolds with almost maximal symmetry rank
	\item Rational ellipticity of nonnegatively curved manifolds with torus actions
	\item Torus orbifolds and slice-maximal actions in nonnegative curvature 
	\item Curvature decay rates on open manifolds with nonnegative curvature 
	\item Soul Theorem generalizations for noncompact nonnegative curvature
	\item Nonnegatively curved manifolds with circle actions and knot characterization 
	\item Cohomogeneity one manifolds with nonnegative curvature
	\item Nonnegative curvature and Alexandrov geometry comparisons
	\item Nonnegative sectional curvature on exotic spheres and diffeomorphism types
	\item Metric classification of nonnegatively curved manifolds in low dimensions
	\item Nonnegative curvature and the Bott conjecture
\end{enumerate}

\paragraph{4. Positive Sectional Curvature and Sphere Theorems}

\begin{enumerate}
	\setcounter{enumi}{60}
	\item Differentiable sphere theorems under positive curvature conditions
	\item Pinching constants and topological implications for positively curved manifolds
	\item Positively curved manifolds with large symmetry groups
	\item Exotic spheres with positive curvature existence problems
	\item Almost positive curvature and its topological restrictions
	\item Quasipositive curvature and symmetry rank bounds 
	\item Positive curvature on homogeneous spaces and classification problems
	\item Eschenburg spaces and their curvature properties
	\item Biquotients with positive curvature and new examples
	\item Positive curvature on 7-manifolds and Aloff-Wallach spaces
	\item Rigidity theorems for positively curved manifolds with symmetry
	\item Grove's symmetry program and classification of positively curved manifolds
	\item Almost nonnegative curvature and Gromov-Hausdorff limits
	\item Positive sectional curvature and the Hopf conjecture
	\item Weyl curvature and positive sectional curvature relations
	\item Positive curvature on orbifolds and resolution of singularities
	\item Hopf's conjecture on positively curved manifolds
	\item Positive curvature and minimal volume estimates
	\item Positive curvature on sphere bundles and bundle constructions
	\item Frankel's theorem and its generalizations
\end{enumerate}

\paragraph{5. Ricci Curvature and Geometric Flows}
\begin{enumerate}
	\setcounter{enumi}{80}
	\item Ricci solitons on Riemannian hypersurfaces in Lorentzian manifolds 
	\item Almost Ricci solitons and their classification on hypersurfaces 
	\item Yamabe solitons on Riemannian hypersurfaces 
	\item Ricci flow with surgeries and long-time existence
	\item Ancient solutions to Ricci flow and their classification
	\item Ricci flow on 4-manifolds and singularity formation
	\item Kähler-Ricci flow and stability of Fano manifolds
	\item Harmonic-Ricci flow and fixed points (harmonic-Einstein metrics) 
	\item Ricci flow on manifolds with boundary and free boundary problems
	\item Convergence of Ricci flow under positive curvature conditions
	\item Ricci limit spaces and their geometric structure
	\item Colding's volume convergence theorems and applications
	\item Ricci curvature and fundamental group growth estimates
	\item Almost Einstein manifolds and rigidity
	\item Ricci curvature and eigenvalue estimates on Riemannian manifolds
	\item Gradient Ricci solitons with nonnegative curvature
	\item Steady Ricci solitons and their asymptotic geometry
	\item Expanding Ricci solitons and cone structures
	\item Shrinking Ricci solitons and classification in low dimensions
	\item Ricci curvature and isoperimetric inequalities
\end{enumerate}
\paragraph{6. Comparison Geometry and Synthetic Curvature Bounds}

\begin{enumerate}
	\setcounter{enumi}{100}
	\item Synthetic Ricci curvature bounds via optimal transport on Riemannian manifolds 
	\item CD(K,N) spaces and Riemannian structure recovery
	\item RCD spaces and their geometric analysis
	\item Unified curvature bounds for Riemannian and sub-Riemannian geometry 
	\item Toponogov-type theorems in Alexandrov geometry with curvature bounds
	\item Bishop-Gromov volume comparison in singular settings
	\item Cheeger-Gromoll splitting theorem generalizations
	\item Laplacian comparison theorems under synthetic Ricci bounds
	\item Sobolev inequalities on RCD spaces and sharp constants
	\item Heat kernel estimates on metric measure spaces with curvature bounds
	\item Bakry-Émery Ricci curvature and applications to diffusion processes
	\item Curvature-dimension conditions for weighted Riemannian manifolds
	\item Metric measure spaces with Riemannian Ricci curvature from below
	\item Gromov-Hausdorff convergence and stability of curvature bounds
	\item Collapsing theory under sectional curvature bounds
	\item Marginally trapped surfaces and comparison geometry in Lorentzian setting 
	\item Generalized Robertson-Walker spacetimes and curvature characterization 
	\item Comparison theorems for submanifolds in warped products
	\item Heintze-Karcher inequality and generalizations
	\item Volume growth estimates on manifolds with nonnegative Ricci curvature
\end{enumerate}
\paragraph{7. Geometric Analysis and PDEs on Manifolds}
\begin{enumerate}
	\setcounter{enumi}{120}
	\item Blow-up mechanisms for evolution equations on noncompact manifolds 
	\item Existence of global solutions to PDEs on complete Riemannian manifolds
	\item Yamabe problem on manifolds with boundary
	\item Prescribed scalar curvature problem on spheres and manifolds
	\item Nirenberg problem on CR manifolds
	\item Q-curvature and Paneitz operator on 4-manifolds
	\item Critical metrics and Bach-flat manifolds
	\item Obata-type rigidity theorems for eigenvalue problems
	\item Steklov eigenvalue problem on Riemannian manifolds with boundary
	\item Isoperimetric inequalities and spectral geometry
	\item Reaction-diffusion equations on evolving manifolds 
	\item Singularity formation in geometric PDEs on noncompact manifolds 
	\item Liouville theorems for harmonic functions on manifolds with curvature bounds
	\item Elliptic systems on Riemannian manifolds
	\item Parabolic equations on manifolds with Ricci flow background
	\item p-Laplacian eigenvalue problems on Riemannian manifolds
	\item Bifurcation theory for geometric PDEs on symmetric spaces
	\item Moser-Trudinger inequalities on compact Riemannian manifolds
	\item Logarithmic Sobolev inequalities and curvature
	\item Geometric variational problems with symmetry constraints
\end{enumerate}
\paragraph{8. Special Structures: Kähler, Quaternion-Kähler, and Spin Geometry}
\begin{enumerate}
	\setcounter{enumi}{140}
	\item Kähler manifolds with positive scalar curvature and harmonic map applications 
	\item Ricci-flat Kähler metrics and Calabi-Yau manifolds
	\item Kähler-Einstein metrics on Fano manifolds
	\item cscK metrics and stability in Kähler geometry
	\item Quaternion-Kähler manifolds with positive scalar curvature and Wolf spaces
	\item Hyperkähler metrics and their deformations
	\item Spin geometry and Dirac operators on manifolds with special holonomy
	\item Positive mass theorem and spin geometry applications
	\item Gromov-Lawson conjecture and metrics of positive scalar curvature
	\item Killing spinors and Einstein manifolds
	\item Sasaki-Einstein manifolds and AdS/CFT correspondence
	\item 3-Sasakian structures and their automorphism groups
	\item Nearly Kähler manifolds and their classification in dimension 6
	\item Calabi-Yau with torsion and heterotic string compactifications
	\item G2 and Spin(7) holonomy manifolds and their construction
	\item Special Lagrangian submanifolds in Calabi-Yau manifolds
	\item Mirror symmetry and special geometry
	\item Hodge theory on non-Kähler manifolds
	\item Balanced metrics and Fu-Yau equations on complex manifolds
	\item SKT metrics and pluriclosed flow
\end{enumerate}
\paragraph{9. Homogeneous Spaces and Lie Groups}
\begin{enumerate}
	\setcounter{enumi}{160}
	\item Left-invariant pseudo-Riemannian metrics on Lie groups 
	\item Geodesic completeness of Lorentzian metrics on Lie groups 
	\item Homogeneous Ricci solitons and their classification
	\item Naturally reductive spaces and their curvature properties
	\item Symmetric spaces of noncompact type and harmonic map rigidity
	\item Hermitian symmetric spaces and bounded domain geometry
	\item Curvature homogeneous manifolds and their classification
	\item Locally homogeneous spaces with special holonomy
	\item Nilpotent Lie groups with left-invariant metrics and Ricci flow
	\item Solvmanifolds with Einstein metrics
	\item Flag manifolds and homogeneous Kähler-Einstein metrics
	\item Weakly symmetric spaces and commutative diagram geometry
	\item Geodesic orbit manifolds and their classification
	\item Eight-dimensional Hermitian Lie groups conformally foliated by minimal leaves 
	\item SU(2)×SU(2)-leaves in Hermitian Lie group foliations 
	\item Homogeneous structures and curvature on reductive spaces
	\item Aloff-Wallach spaces and their curvature properties
	\item Wallach spaces and positive curvature examples
	\item Berger spheres and their geometric properties
	\item Homogeneous geodesics in Riemannian homogeneous spaces
\end{enumerate}
\paragraph{10. Sub-Riemannian Geometry and Generalizations}
\begin{enumerate}
	\setcounter{enumi}{180}
	\item Synthetic curvature bounds unifying Riemannian and sub-Riemannian geometry 
	\item Ricci curvature in sub-Riemannian manifolds via optimal transport 
	\item Heisenberg group geometry and curvature-dimension conditions 
	\item Carnot groups and their metric geometry
	\item Contact metric manifolds and critical compatible metrics 
	\item Biconservative PMCV surfaces in Robertson-Walker spaces 
	\item CR manifolds and pseudohermitian geometry
	\item Sub-Riemannian geodesics and Hamiltonian dynamics
	\item Hörmander's condition and hypoelliptic operators on manifolds
	\item Sub-Laplacian eigenvalue estimates on CR manifolds
	\item Sasakian geometry and its sub-Riemannian aspects
	\item Quasi-contact structures and curvature properties
	\item Sub-Riemannian isoperimetric inequalities and applications
	\item Gromov's filling volume in sub-Riemannian geometry
	\item Heat kernel asymptotics on sub-Riemannian manifolds
	\item Tangent cones and metric measure spaces in sub-Riemannian setting
	\item Sub-Riemannian manifolds with symmetries and reduction
	\item Abnormal geodesics in sub-Riemannian geometry
	\item Sub-Riemannian structures on Lie groups and optimal control
	\item Infinite-dimensional Riemannian geometry and shape analysis applications 
\end{enumerate}

\end{document}